\documentclass{article}
\usepackage{ijcai21}

\usepackage{times}

\usepackage{soul}
\usepackage{url}
\usepackage[utf8]{inputenc}
\usepackage[small]{caption}
\usepackage{graphicx}
\usepackage{amsmath}
\usepackage{booktabs}
\usepackage{mathrsfs}
\usepackage{makecell}
\usepackage{subfigure}
\usepackage{booktabs}
\usepackage{color}
\usepackage{multirow}
\usepackage{graphics}
\usepackage{lipsum}
\usepackage{amssymb}
\usepackage{dsfont}
\usepackage{bbm}
\usepackage{xcolor}
\usepackage{colortbl}
\usepackage[pagebackref=true,breaklinks=true,colorlinks,bookmarks=false]{hyperref}

\def \BA{\mathbf{A}}

\def \BX{\mathbf{X}}
\def \BY{\mathbf{Y}}

\def \BD{\mathbf{D}}
\def \BE{\mathbf{E}}

\def \BN{\mathbf{N}}
\def \BG{\mathbf{G}}

\def \R{{\mathbb{R}}}
\newcommand{\pare}[1]{{(#1)}}  % parethese sth, with extra {} added, used in a
\title{On Self-Distilling Graph Neural Network}

\author{
Yuzhao Chen,\textsuperscript{\rm 1,}\textsuperscript{\rm 2}\and
Yatao Bian,\textsuperscript{\rm 2}\and
Xi Xiao,\textsuperscript{\rm 1}\and
Yu Rong,\textsuperscript{\rm 2}\and
Tingyang Xu,\textsuperscript{\rm 2}\and
Junzhou Huang \textsuperscript{\rm 2,}\textsuperscript{\rm 3}\\
\affiliations
\textsuperscript{\rm 1}Tsinghua Shenzhen International Graduate School, Tsinghua University, Shenzhen, China 518057, \\
\textsuperscript{\rm 2}Tencent AI Lab, Shenzhen, China 518057, \\
\textsuperscript{\rm 3}University of Texas at Arlington, Arlington, TX 76019,
}

\begin{document}

\maketitle

\begin{abstract}
Recently, the teacher-student knowledge distillation framework has demonstrated its potential in training Graph Neural Networks (GNNs). 
However, due to the difficulty of training over-parameterized GNN models, one may not easily obtain a satisfactory teacher model for distillation. Furthermore, the inefficient training process of teacher-student knowledge distillation also impedes its applications in GNN models. 
In this paper, we propose the \emph{first} teacher-free knowledge distillation method for GNNs, termed GNN Self-Distillation (GNN-SD),
that serves as a drop-in replacement of the standard training process.
The method is built upon the proposed neighborhood discrepancy rate (NDR), which quantifies the non-smoothness of the embedded graph in an efficient way. 
Based on this metric, we propose the adaptive discrepancy retaining (ADR) regularizer to empower the transferability of knowledge that maintains high neighborhood discrepancy across GNN layers. 
We also summarize a generic GNN-SD framework that could be exploited to induce other distillation strategies.
Experiments further prove the effectiveness and generalization of  our approach, as it brings:  
1) state-of-the-art GNN distillation performance with less training cost, 
2) consistent and considerable performance enhancement for various popular backbones. 
\end{abstract}

\section{Introduction}
\label{sec:intro}

Knowledge Distillation (KD) has demonstrated its effectiveness in boosting compact neural networks. 
Yet, most of the KD researches focus on Convolutional Neural Networks (CNNs) with regular data as input instances, while little attention has been devoted to Graph Neural Networks (GNNs) that are capable of processing irregular data. 
A significant discrepancy  is that GNNs involve the topological information into the updating of feature embeddings across network layers, which is not taken into account in the existing KD schemes,
restricting their potential  extensions to GNNs. 

A recent work, termed LSP~\cite{yang2020distilling},  proposed to combine KD with GNNs by transferring the local structure, which is modeled as  the distribution of the similarity of connected node pairs, from a pre-trained teacher GNN to a light-weight student GNN.
% with fewer layers or lower feature dimensions.
However, there exists a major concern on the selection of qualified teacher GNN models.
On the one hand, it's likely to cause performance degradation once improper teacher networks are selected~\cite{tian2019contrastive}.
On the another hand, the performance of GNNs is not always indicated by their model capacity due to the issues of over-smoothing~\cite{li2018deeper} and over-fitting~\cite{rong2019dropedge}, which have caused obstacles to train over-parameterized and powerful GNNs. 
As a result, every time encountering  a new learning task, one may spend substantial efforts in searching for a qualified GNN architecture to work as a teacher model, 
and thus the generalization ability of this method remains a challenge.
Another barrier of the adopted conventional teacher-student framework is the inefficiency in the training process. Such distillation pipeline usually requires two steps: first,  pre-training a relatively heavy model, and second,  transferring the forward predictions (or transformed features) of the teacher model to the student model.
With the assistance of the  teacher model, the training cost would tremendously increase more than twice than an ordinary training procedure. 

In this work, we resort to cope with these issues via the self-distillation techniques (or termed teacher-free distillation), which perform knowledge extraction and transfer between layers of a single network without the assistance from auxiliary models~\cite{zhang2019your,hou2019learning}.
Our work provides the first dedicated self-distillation approach designed for generic GNNs, named GNN-SD.
The core ingredient of GNN-SD is motivated by the mentioned challenge of over-smoothing, which occurs when GNNs go deeper and lead the node features to lose their discriminative power. 
Intuitively, one may avoid such dissatisfied cases by pushing node embeddings in deep layers to be distinguishable from their neighbors, which is exactly the property possessed by shallow GNN layers.

To this end, we first present the \emph{Neighborhood Discrepancy Rate} (NDR) to serve as an approximate metric in quantifying the non-smoothness of the embedded graph at each GNN layer.
Under such knowledge refined by NDR, 
we propose to perform  knowledge self-distillation by  an adaptive discrepancy retaining (ADR) regularizer. The ADR regularizer adaptively selects the target knowledge contained in shallow layers as the supervision signal and retains it to deeper layers. 
Furthermore, we summarize a generic GNN-SD framework that could be exploited to derive other distillation strategies. 
As an instance, we extend GNN-SD to involve another knowledge source of \emph{compact graph embedding} for better matching the requirements of graph classification tasks. 
% The overall framework is depicted in Figure ~\ref{fig:framework}.
In a nutshell, our main contributions are:
\begin{itemize}
    \item We present GNN-SD, to our knowledge, the first generic framework designed for distilling the graph neural networks with no assistance from extra teacher models.
    It serves as a drop-in replacement of the standard training process to improve the  training dynamics. 
    \item We introduce a simple yet efficient  metric of NDR to refine the knowledge from each GNN layer. Based on it, the ADR regularizer is proposed to  empower the adaptive knowledge  transfer inside  a   single GNN model. 
    
    \item We validate the effectiveness and generalization ability of our GNN-SD by conducting experiments on multiple popular GNN models, yielding the state-of-the-art distillation result and consistent performance improvement against baselines.
\end{itemize}

\section{Related Work}\label{sec:related}

\begin{figure*} [t]
  \centering
  \includegraphics[width=1.0\linewidth]{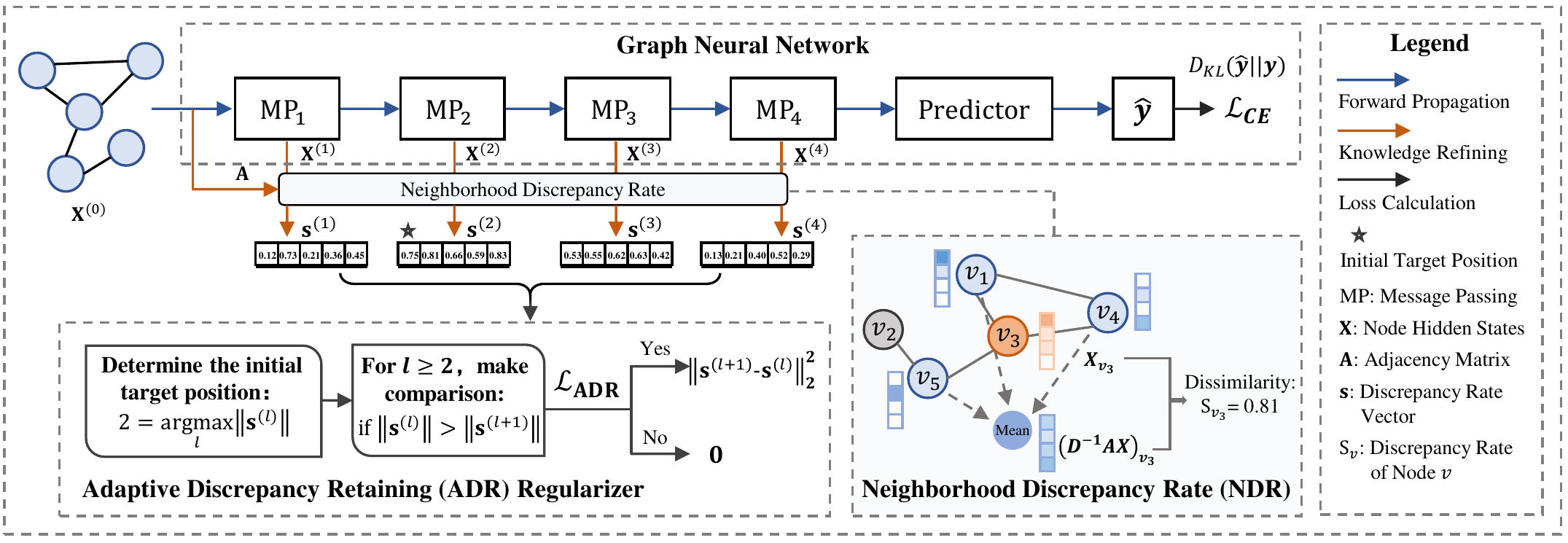}
  \vspace{-4mm}
  \caption{The schemata of the distillation strategy of adaptive discrepancy retaining. 
    A four-layer GNN is adopted for the  illustration.
 }
 \label{fig:framework} 
 \vspace{-2mm}
\end{figure*}

\paragraph{Graph Neural Network} 
Recently, Graph Neural Networks (GNNs), which propose to perform message passing across nodes in the graph and updating their representation, has achieved great success on various tasks with irregular data, such as node classification, protein property prediction to name a few. 
Working as a crucial tool for graph representation learning, however, these models encounter the challenge of over-smoothing. 
It says that the representations of the nodes in GNNs would converge to a stationary point and become indistinguishable from each other when the number of layers in GNNs increases.  
This phenomenon limits the depth of the GNNs and thus hinders their representation power. 

One solution to alleviate this problem is to design network architectures that can better memorize and utilize the initial node features.
Representative papers includes GCN~\cite{kipf2016semi}, JKNet~\cite{xu2018representation} DeeperGCN~\cite{li2020deepergcn},  GCNII~\cite{chen2020simple}, etc.
On the other hand, methods like DropEdge~\cite{rong2019dropedge} and AdaEdge~\cite{chen2020measuring} have proposed solutions from the view of conducting data augmentation.

In this paper, we design a distillation approach tailored for GNNs, which also provides a feasible solution to this problem.
Somewhat related, \citeauthor{chen2020measuring}~\shortcite{chen2020measuring} proposed a regularizer to the training loss, which simply forces the nodes in the  last GNN layer to obtain a large distance between remote nodes and their neighboring nodes. However, it can only obtain a slight performance improvement.

\paragraph{Teacher-Student Knowledge Distillation}
Knowledge distillation~\cite{KD}, aims at transferring the knowledge hidden in the target network (i.e. teacher model) into the online network (i.e. student model) that is typically light-weight, so that the student achieves better performance compared with the one trained in an ordinary way.
Generally, there exist two technical routes for KD. 
The first one is closely related  to label smoothing~\cite{yuan2020revisiting}, which utilizes the output distribution of the teacher model to serve as a smooth label for training the student. 
Another line of research  is termed as feature distillation~\cite{romero2014fitnets,zagoruyko2016paying,kim2018paraphrasing}, 
which exploits the semantic information contained in the intermediate representations. 
As summarized in~\cite{heo2019comprehensive}, with different concerning knowledge to distill, these methods can be distinguished by the formulation of feature transformation and knowledge matching loss function.

Recently, ~\citeauthor{yang2020distilling}~\shortcite{yang2020distilling} studied the teacher-student distillation methods in  training GNNs. They extract the knowledge of local graph structure based on the similarity of connected node pairs 
from the teacher model and student model, then perform distillation by forcing the student model to match such knowledge.
However, the performance improvement  resulted from the teacher-student distillation framework does not come with a free price, as discussed in Section~\ref{sec:intro}. 

\paragraph{Self-Distillation} 
For addressing the issues of the teacher-student framework, a new research area termed  teacher-free distillation, or self-distillation, attracts a surge of attention recently.
Throughout this work, we refer this notion to the KD techniques that perform knowledge refining and transfer between network layers inside a single model.
In this way, the distillation learning could be conducted with a single forward propagation in each training iteration. 
BYOT~\cite{zhang2019your} proposed the first  self-distillation method. They consider that the  teacher and student are composed in the same networks since the deeper part of the networks can extract more semantic information than the shallow one. Naturally, they manage to distill  feature representations as well as the smooth label  from deeper layers into the shallow layers.
Similarly, \citeauthor{hou2019learning}~\shortcite{hou2019learning} proposed to distill the attention feature maps from the deep layers to shallow ones for lane detection. 
However, these methods focus on the application of CNNs, neglecting the usage of graph topology information, and thus restricting their potential extension to GNNs.

\section{GNN Self-Distillation}\label{sec:na}

A straightforward solution to perform self-distillation for training GNNs is to supervise the hidden states of shallow layers by the ones of deep layers as the target, as the scheme proposed in BYOT~\cite{zhang2019your}. 
However, we empirically find that such a strategy leads to performance degradation. 
It's on the one hand attributed   to the  neglection of the graph topological information. On the other hand,  it's too burdensome for shallow GNN layers to match their outputs to  such unrefined knowledge and eventually leads to over-regularizing.
Furthermore, 
such a simple solution requires fixed representation dimensionalities across network layers,  which limits its applications to generic GNN models.
In the following sections, we first introduce the key ingredient of our GNN Self-Distillation (GNN-SD). Then we summarize a unified GNN-SD framework that is well extendable to other distillation variants.

\paragraph{Notation} 
Throughout this work, a graph is represented  as $\mathcal{G}=\{\mathcal{V}, \mathcal{E}, \BA\}$, 
where $\mathcal{V}$ is  vertex set that has $N$ nodes with $d$-dimension features of $\BX\in \mathbb{R}^{N\times d}$,
edge set $\mathcal{E}$ of size $M$  is encoded with edge features of  $\BE$,
and $\BA\in \mathbb{R}^{N\times N}$ is the adjacency matrix.
The node degree matrix is given by $\BD=\mathrm{diag}(\BA \mathbf{1}_{N})$.
Node hidden states of $l$-th GNN layer is denoted as $\BX^\pare{l}$, and the initial hidden states $\BX^\pare{0}$ is usually set as the node intrinsic features $\BX$.
Given a node $v$, its connected-neighbors are denoted as $\mathcal{N}_v$. For a matrix $\BX$, $\BX_{i\cdot}$ denotes its $i$-th row and $\BX_{\cdot j}$ denotes its $j$-th column.

\subsection{Adaptive Discrepancy Retaining}
\label{sec:ndr}

Since the well-known over-smoothing issue of GNNs occurs when the input graph data flows into deep layers,
an inspired insight is that we can utilize the property of non-smoothness
in shallow GNN layers, and  distill such knowledge into  deep ones. 
In this way, the model is self-guided to retain non-smoothness from the initial embedded graph to the final embedded output.
The remained  questions are: how to refine the desired knowledge from the fine-grained node embeddings? and what is a proper knowledge transfer strategy?

\paragraph{Neighborhood Discrepancy Rate}
To answer the first question,
we  introduce  the module of Neighborhood Discrepancy Rate (NDR), 
which is used   to  quantify the non-smoothness of each GNN layer.

It's proved that 
nodes in a graph component converge to the same stationary point while 
iteratively updating the message passing process of GNNs, 
and hence the hidden states of connected-nodes become indistinguishable~\cite{li2018deeper}. It implies that, given a node $v$ in the graph,
\begin{equation}\label{eq:same-stationary}
\sum\nolimits_{c\in\mathcal{N}_v}\Vert\BX_{v\cdot}^\pare{l}-\BX_{c\cdot}^\pare{l}\Vert_p <\epsilon, \text{if}\;l\to\infty, \forall\epsilon>0.
\end{equation}
As a result, it leads to the issue of over-smoothing and further hurts the accuracy of node classification.
Centering around this conclusion, it is a natural choice to use the pair-wise metric (and the resulting distance matrix) to define the local non-smoothness of the embedded graph. However, such fine-grained knowledge might still cause over-regularization for GNNs trained under teacher-free distillation. 
By introducing the following proposition (details deferred to Appendix \ref{proof-1}), one can easily derive from formula (\ref{eq:same-stationary}) that, 
$\Vert\BX_{v\cdot}^\pare{l} - (\BD^{-1}\BA\BX)_{v\cdot}^\pare{l}\Vert_p<\epsilon$ as layer $l$ goes to infinity.

\vspace{1mm}
\noindent\textbf{Proposition 1.} \textit{Suppose $d_1(\BX, \mathcal{G})$ calculates the $L_p$-norm of the difference of each central node and their aggregated neighbors in the graph $\mathcal{G}$, and the pair-wise distance metric $d_2(\BX, \mathcal{G})$, on the other hand,  computes the difference at the level of node pairs. Then, the inequality holds: $d_2(\BX, \mathcal{G}) \geq d_1(\BX, \mathcal{G})$.}
\vspace{0.5mm}
\\
We leverage this property to refine the knowledge from a higher level. 
Specifically, given a central node $v$ in layer $l$, we first obtain the  aggregation of its adjacent nodes to work as the virtual node that indicates its overall neighborhood, $\BN_v^\pare{l} = (\BD^{-1}\BA\BX^\pare{l})_{v\cdot}$.
For excluding the effect of embeddings' magnitude, we use the cosine similarity between the embeddings of central node $\BX^\pare{l}_{v\cdot}$ and virtual node $\BN^\pare{l}_{v}$ to calculate their affinity, and transform it into a distance metric,
\begin{equation}\label{eq:node-smooth-i}
S_v^\pare{l} = 1 - \frac{{\BX^\pare{l}_{v\cdot}}(\BA \BX^\pare{l})_{v\cdot}^\text{T}}{||\BX^\pare{l}_{v\cdot}||_2\cdot||(\BA\BX^\pare{l})_{v\cdot}||_2}, v=1,...,N,
\end{equation}
Note that it's not needed to perform  the inverse matrix multiplication of node degrees due to the  normalization conducted by cosine similarity metric.
The defined $S_v^\pare{l}$ of all nodes 
compose the neighborhood discrepancy rate  of layer $l$:
\begin{equation}\label{eq:node-smooth}
\mathbf{s}^\pare{l}  = (S_1^\pare{l}, ..., S_{N}^\pare{l}).
\end{equation}

Compared with the pair-wise metric, the NDR extracts neighbor-wise non-smoothness, which is easier to transfer and prevents over-regularizing by self-distillation.
Moreover, it can be easily implemented with two consecutive matrix multiplication operations, enjoying a significant computational advantage. The NDR also possesses better flexibility to model local non-smoothness of the graph, since pair-wise metrics can not be naturally applied together with layer-wise sampling techniques~\cite{chen2018fastgcn,huang2018adaptive}.

Specially, for the task of node classification, there is another reasonable formulation of the virtual neighboring node. 
That is, taking  node labels into account,  
$\BN_v^\pare{l} = (\BD^{-1}\BA^{'}\BX^\pare{l})_{v\cdot}$, where $\BA^{'}=\BA\odot\BY$ denotes the masked adjacency, $\BY\in\mathbb{R}^{N\times N}$ the binary matrix with entries $\BY_{i,j}$  equal to 1 if node $i$ and $j$ are adjacent and belong to different categories, and $\odot$ the element-wise multiplication operator. 
Then, the NDR  would not count the discrepancy of nodes that are supposed to share high similarity.
For unity and simplicity, we still use the former definition throughout this work.

\paragraph{Strategy for Retaining Neighborhood Discrepancy}
Previous self-distillation methods usually treat the deep representations as the target supervision signals, since they are considered to contain more semantic information.
However, we found that such a strategy is not optimal, sometimes even detrimental for training GNNs (refer to Appendix \ref{append:two-scheme} for details). 
The rationale behind our design of distilling neighborhood discrepancy is to retain the non-smoothness, which is extracted as the knowledge by NDR, from shallow GNN layers to the deep ones.
In details, we design the following guidelines (refer to Appendix~\ref{append:explamar} for more analysis) for the self-distillation learning, which formulate our adaptive discrepancy retaining (ADR) regularizer:
\begin{itemize}
\item The noise in shallow network layers might cause the calculated  NDR of the first few embedded graphs to be inaccurate, thus the initial supervision target is adaptively determined by the magnitude of the calculated NDR.
\item For facilitating the distillation learning, the knowledge transfer should be progressive. Hence, the ADR loss is computed by matching the NDR of deep layer (online layer) to the target one of its previous layer (target layer).
\item Explicit and adaptive teacher selection is performed, i.e the ADR regularizes the GNN only when the magnitude of NDR of the target layer is larger than the online layer. 
\item Considering that the nodes in regions of different connected densities have different rates of becoming over-smoothing~\cite{li2018deeper}, the matching loss can be weighted by the normalized node degrees to emphasize such a difference.
\end{itemize}
As a result, the final ADR regularizer is defined as:
\begin{equation}\label{eq:node-level-sd}
\mathcal{L}_{\mathrm{N}} = \sum_{l=l^\star,...,L-1}\mathbbm{1}(\Vert\mathbf{s}^\pare{l}\Vert\textgreater \Vert\mathbf{s}^\pare{l+1}\Vert)d^2(\mathbf{s}^\pare{l+1}, \mathbf{s}^\pare{l}),
\end{equation}
where the indicator function $\mathbbm{1}(\cdot)$ performs the teacher selection, $l^\star = \mathrm{argmax}_{k}\{\Vert\mathbf{s}^\pare{k}\Vert | k\in\{1,...,L-1\}\}$ determine the position of initial supervision target, and $d^2(\mathbf{s}^\pare{l+1}, \mathbf{s}^\pare{l})=\Vert\BD(\mathbf{s}^\pare{l+1}-\mathrm{SG}(\mathbf{s}^\pare{l}))^\text{T}\Vert_2^2$ is the degree-weighted mean squared error function that calculates the knowledge matching loss. 
Here $\mathrm{SG}(\cdot)$ denotes the Stop Gradient operation, meaning that the gradient of   the target NDR  tensor is detached in the implementation, for serving as a supervision signal.
The approach is depicted in Figure \ref{fig:framework}.

In addition, we analytically demonstrate that the proposed notion of  discrepancy retaining  can be comprehended from the perspective of information theory. This is analogous to the concept in~\cite{ahn2019variational}. Specifically, the retaining of neighborhood discrepancy rate  encourages the online layer to share high mutual information with the target layer, as illustrated in the following proposition ($I$ stands for mutual information and $H$ denotes the entropy, details are deferred to Appendix~\ref{proof-2}).

\noindent\textbf{Proposition 2.} \textit{The optimization of the ADR loss increase the lower bound of the mutual information between the target NDR and the online one. That is, the inequality holds:} $I(\mathbf{s}^\pare{l}, \mathbf{s}^{\pare{l+1}})\geq H(\mathbf{s}^\pare{l}) -  \mathbbm{E}_{\mathbf{s}^\pare{l},\mathbf{s}^\pare{l+1}}[\Vert \BD (\mathbf{s}^\pare{l+1}-\mathbf{s}^\pare{l})^\text{T} \Vert_2^2]$.

\subsection{Generic GNN-SD Framework} 
Generally, by refining and transferring  compact and informative  knowledge   between  layers, self-distillation on GNNs can be  summarized as the learning of the additional mapping,
\begin{equation}\label{eq:Mapping-distill}
\mathcal{M}_{\mathrm{SD}}^{g, L}: \mathcal{T}_{s}(\mathcal{C}_{s}(\mathcal{G}, P_s))\to \mathcal{T}_{t}(\mathcal{C}_{t}(\mathcal{G}, P_t)),
\end{equation}
where  $P\in \{1,...,L\}$ is the layer position to extract knowledge from the network,
$\mathcal{C}$ denotes the granularity (or coarseness) of the embedded graph, 
$\mathcal{T}$ represents the specific transformation applied to the chosen embeddings, 
and the subscripts of $s$ and $t$ denote the identity of student (to simulate) and teacher (to transfer), respectively. 

Naturally, the combinations of  different granularities and transformation functions  lead to various distilled knowledge. 
As an instance, we show here to involve another knowledge source of the full-graph embedding, and provide further discussions in Appendix~\ref{append:framework} for completeness.

Considering the scenario of graph classification, where GNNs might focus more on obtaining meaningful embedding of the entire graph than individual nodes, 
the full-graph embedding could be the well-suited knowledge, since it provides a global view of the embedded graph (while the NDR captures the local property),
\begin{equation}\label{ref:graph_readout}
    \mathcal{C(\mathcal{G},P)} := \BG^\pare{P}=\mathrm{Readout}_{v\in \mathcal{G}}(\BX^\pare{P}_{v\cdot}),
\end{equation}
where $\mathrm{Readout}$ is a permutation invariant operator that aggregates embedded nodes to a single embedding vector. 
In contrast to the fine-grained node features, the coarse-grained graph embedding is sufficiently compact so as we can
simply use the identity function 
to preserve the transferred knowledge. Hence the target mapping is:
\begin{equation}\label{eq:Mapping-graph}
\mathcal{M}_{\mathrm{graph}}^{g, L}: \mathbf{\mathds{1}}(\BG^\pare{l+1})\to \mathbf{\mathds{1}}(\BG^\pare{l}).
\end{equation}
It can be learned by optimizing the graph-level distilling loss:
\begin{equation}\label{eq:graph-level-sd}
L_{\mathrm{G}} = \sum\nolimits_{l=1,...,L-1}||\BG^\pare{l+1}-\mathrm{SG}(\BG^\pare{l})||_2^2.
\end{equation}
In this way, GNN-SD extends the utilization of mixed knowledge sources over different granularities of the embedded graph. 

\begin{figure*}[t]
  \centering
  \includegraphics[width=0.83\linewidth]{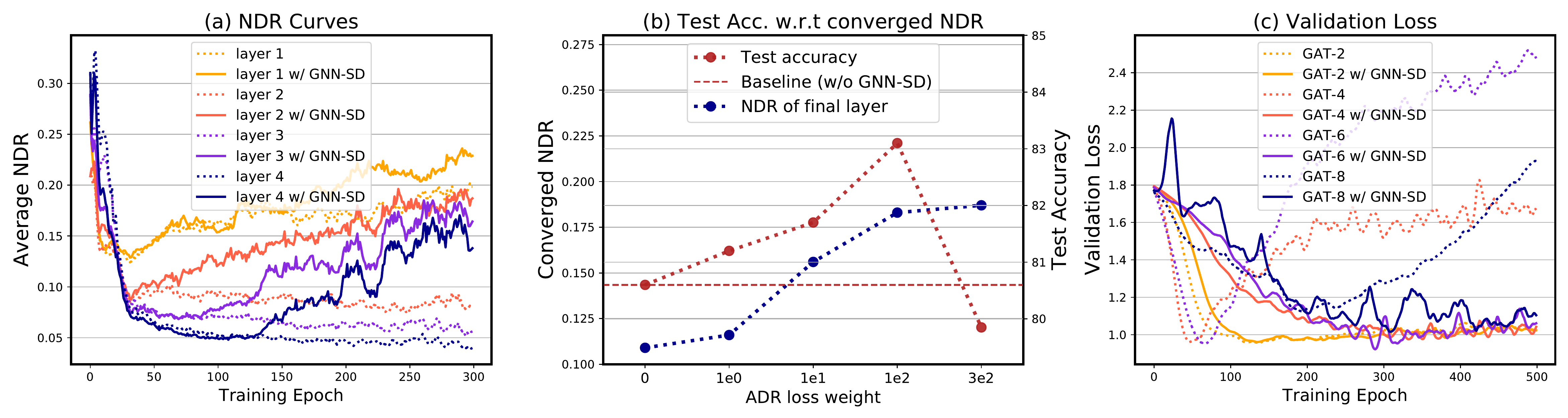}%\vspace{-2mm}
  \vspace{-3mm}
  \caption{(a) Comparison of NDR between training w/ (solid line) and w/o (dotted line) imposing ADR regularizer. Smaller value means suffering from more over-smoothing. (b) The correlation between test accuracy and converged NDR of the final layer, controlled by the loss weights of ADR regularizer. (c) Comparison of validation loss curves. GAT-$n$ denotes GAT model with $n$ hidden layers.}
 \label{fig:analysis-rnd} 
 \vspace{-1mm}
\end{figure*}

\paragraph{Overall Loss}
The total loss function is formulated as:
\begin{equation}\label{eq:total-object}
\mathcal{L}_{\mathrm{T}} = \mathrm{CE}(g(\BX^\pare{L}), y) + \alpha \mathcal{L}_{\mathrm{L}} + \beta \mathcal{L}_{\mathrm{N}} + \gamma \mathcal{L}_{\mathrm{G}}.
\end{equation}
The first term calculates the basic cross entropy loss between the final predicted distribution and the ground-truth label $y$. The second term, borrowing from~\cite{zhang2019your}, regularizes the intermediate logits generated by intermediate layers to mimic the final predicted distribution for accelerating the training and improving the capacity of shallow layers. 
The remaining terms are defined in  
Eq.(\ref{eq:node-level-sd}) and
Eq.(\ref{eq:graph-level-sd}).
$\alpha$, $\beta$, and $\gamma$ are the hyper-parameters that balance the supervision of the distilling objectives and target label.

\section{Experiments}\label{sec:exp}
\subsection{Exploring Analysis of Discrepancy Retaining}
We first conduct exploring analysis to investigate how the ADR regularizer  helps improve the training dynamics of GNNs.  
Hyper-parameters of $\alpha$ and $\gamma$ are fixed to $0$.

In Figure ~\ref{fig:analysis-rnd}(a), we show the comparison of NDR curves 
of each layer in a $4$-layer GraphSage~\cite{hamilton2017inductive}. 
In line with the expectations, 
the neighborhood discrepancy in a GNN model drops when the layer grows, 
and  the final layer (in dark blue, dotted line) approaches  a close-to-zero   value of NDR,
which indicates that the output node embeddings become indistinguishable with their neighbors. 
Conversely, the GNN trained with GNN-SD preserves higher discrepancy over shallow to deep layers,
and even reveals an increasing tendency as the training progresses.
It might imply that the GNN gradually learns to pull the connected-node embeddings away if they're not of the same class, for obtaining more generalized representations. 
This observation indicates ADR's effect on \emph{alleviating the over-smoothing issue}.
We also provide  related examples in Appendix \ref{append:explamar} that motivate the development of ADR regularizer.

In Figure ~\ref{fig:analysis-rnd}(b), we study the correlation of model performance and the converged NDR:
as the ADR loss weight increases in a reasonable range (e.g. from $1e0$ to $1e2$),  the NDR increases and the performance gain improves (from $0.6\%$ to $2.5\%$).
The shown \emph{positive correlation between test accuracy and NDR} verifies the rationality of the NDR metric and the distillation strategy as well.
Notably, there also exists a trade-off in determining the optimal loss weight.
If an over-estimated weight ($1e3$) was assigned, the knowledge transfer task would become too burdensome  for the GNN model to learn the main classification task and hurt the performance.   

Figure ~\ref{fig:analysis-rnd}(c) depicts the  validation loss on GAT~\cite{velivckovic2018graph} backbones with different depths.
The validation loss curves are dramatically pulled down after applying GNN-SD. 
It explains that ADR regularizer also helps GNNs \emph{relieve the issue of over-fitting}, which is known as another tough obstacle in training deep GNNs.

\begin{table}[thbp] %[hth]
\centering
% \vspace{-2mm}
\caption{Performance comparison with other distillation methods. 
The second column indicates the knowledge source, $\BX$ is node features, $\BA$ is the adjacency matrix, $T(\cdot)$ denotes  the teacher model.
}
\vspace{-2mm}
% \vspace{0.5mm}
\resizebox{0.85\linewidth}{!}{
\begin{tabular}{cc|c|c}
\toprule
      $\textbf{Method}$   & $\textbf{Knowledge Source}$    &   $\textbf{F1 Score}$   &        $\textbf{Time}$   \\ 
\midrule
% \midrule
  Teacher ($T$)  &  /   & $97.6$                      &  0.85s              \\ 
    Baseline  &  /    &  $95.7$   & $\textbf{0.62s}$      \\
% \midrule
     AT          &  $T(\BX)$   &  $95.4$        & $\text{1.75s}$      \\
     FitNet      &  $T(\BX)$    &  $95.6$     & $\text{1.99s}$       \\
     LSP         &  $T(\BX)$,$T(\BA)$   &  $96.1$     & $\text{1.90s}$       \\
\midrule
    Baseline  &  /    &   $95.61_{\pm0.20}$ & $\textbf{0.62s}$      \\ 
     AT         &  $\BX$   &   $95.88_{\pm0.25}$       & $\text{0.73s}$      \\
     FitNet          &  $\BX$    &   $95.60_{\pm0.17}$       & $\text{0.95s}$      \\
     BYOT         &  $\BX$   &   $95.81_{\pm0.56}$    &   $\text{0.80s}$     \\
% \midrule
    GNN-SD      & $\BX$,$\BA$    &       $\textbf{96.20}_{\pm0.03}$ &   $\text{0.87s}$  \\
% \hline
\bottomrule
\end{tabular}
}
\label{tab:compCVPR}
\vspace{-2mm}
\end{table}

\subsection{Comparison with KD Methods}\label{sec:vsCVPR}
We compare our method with other distillation methods, including AT~\cite{zagoruyko2016paying}, FitNet~\cite{romero2014fitnets}, BYOT~\cite{zhang2019your} and LSP~\cite{yang2020distilling}. 
We follow \cite{yang2020distilling} to perform the comparisons on the baseline of a 5-layer GAT on the PPI dataset~\cite{zitnik2017predicting}. 
For evaluating AT and FitNet in a teacher-free KD scheme, we follow their papers to perform transformations on node embeddings to get the attention maps and intermediate features, respectively.
We also evaluate BYOT to see the effect of intermediate logits.
The experiments are conducted for 4 runs, and  those under teacher-student framework are cited from~\cite{yang2020distilling}.
For our method, the hyper-parameters of $\alpha$ and $\beta$ are both set to $0.01$ and $\gamma$ is $0$.
Training time (per epoch) of each model is measured on a NVIDIA 2080 Ti GPU.
Results are summarized in Table~\ref{tab:compCVPR}. 
More details are deferred to Appendix~\ref{append:LSP}.
Clearly, GNN-SD obtains significant performance promotion against other self-distillation methods by involving the topological information into the refined knowledge.
GNN-SD also achieves a better performance gain (0.59) on the baseline compared with LSP (0.4),  even when our method does not require the assistance from an extra teacher model. 
In this way, the training cost is greatly saved (40$\%$ training time increase against baseline for GNN-SD v.s. 190$\%$ increase for LSP). Besides, it avoids the process of selecting qualified teachers, bringing much better usability.

\begin{table}[ht] %[hth]
\centering
\caption{Node classification on varying-depths models.  
}
\vspace{-2mm}
% \vspace{0.5mm}
\resizebox{0.95\linewidth}{!}{
\begin{tabular}{cc|cccc}
\toprule
    \multirow{2}{*}{$\textbf{Dataset}$} &    \multirow{2}{*}{$\textbf{Model}$}    &   \multicolumn{4}{c}{$\textbf{Layers}$}  \\
                            &                                &   2   &   4   &   8   & 16   \\
\midrule
    \multirow{4}{*}{Cora}                &       GAT            &   83.2                &   80.1                    &   76.9                &     74.8             \\
                                        &       GAT w/ GNN-SD        &  \textbf{83.7}        &   81.2                &   \textbf{80.1}        &    77.6              \\
                                          &       GraphSage        &  81.3             &   80.3                    &     78.8              &     77.2             \\
                                          &     GraphSage w/ GNN-SD    &  81.7             &    \textbf{81.5}       &    79.8                  &   \textbf{78.2}              \\
\midrule
    \multirow{4}{*}{Citeseer}                &       GAT           &  72.5                 &    70.5               &    65.1               &    64.5              \\
                                        &       GAT w/ GNN-SD       &  72.6                 &    \textbf{71.5}       &    \textbf{68.3}       &    \textbf{66.2}              \\
                                          &       GraphSage         &  72.3                 &    70.7               &    61.7               &    59.2              \\
                                          &     GraphSage w/ GNN-SD    &  \textbf{72.7}        &    71.0               &    64.5               &    61.8              \\
\midrule
    \multirow{4}{*}{Pubmed}                &       GAT              &   79.2            &    78.5                   &    76.6               &    75.6              \\
                                        &       GAT w/ GNN-SD          &   \textbf{79.5}   &    \textbf{79.4}          &    \textbf{78.5}       &    76.6              \\
                                          &       GraphSage      &   78.8            &    77.9                   &    73.8               &    77.2              \\
                                          &     GraphSage w/ GNN-SD    &   79.2            &    79.4                   &    77.6               &   \textbf{78.2}              \\
% \hline
\bottomrule
\end{tabular}
}
\label{tab:various-depth}
\vspace{-3mm}
\end{table}

\subsection{Overall Comparison Results}
\paragraph{Node Classification} Table \ref{tab:various-depth} summarizes the results of GNNs with various depths on the citation classification datasets, including Cora, Citeseer and PubMed~\cite{sen2008collective}. 
We follow the setting of semi-supervised learning, using the fixed 20 nodes per class for training.
The  hyper-parameters of $\gamma$ is fixed to $0$ for node classification, and we determine  $\alpha$ and $\beta$ via  a simple grid search. Details are provided in Appendix \ref{append:node-hyper}. 
It’s observed that GNN-SD consistently improves the test accuracy for all cases.
Generally, GNN-SD yields larger improvement for deeper architecture, as it gains $0.5\%$ average improvement for a two-layer GAT on Cora while achieving $3.2\%$ increase for the 8-layer one.

\paragraph{Graph Classification} Table \ref{tab:table1-mainComparison} summarizes the results of various popular GNNs on the graph kernel classification datasets, including ENZYMES, DD, and PROTEINS in TU dataset~\cite{KKMMN2016}. 
Since there exist no default splittings, each experiment is conducted by 10-fold cross validation with the  splits ratio at  8:1:1 for training, validating and testing.
We  choose five widely-used GNN models, including GCN, GAT, GraphSage, GIN~\cite{xu2018powerful} and GatedGCN~\cite{bresson2017residual}, to work as the evaluation baselines. 
Hyper-parameter settings are deferred to Appendix~\ref{append:graph-hyper}.
Again, one can observe that GNN-SD achieves consistent and considerable performance enhancement against all the baselines.
On classification of PROTEINS, for example,  even  the next to last model trained with GNN-SD (GraphSage, $76.71\%$) outperforms the best model (GAT, $76.36\%$) trained in the ordinary way.  
The results further validate the generalization ability of our self-distilling strategy.
% \vspace{-2mm}

\begin{table}[bht] %[hth]
\centering
% \vspace{-1mm}
\caption{Graph classification on various GNN backbones. 
The top 2 models on each dataset are bolded.}
\vspace{-2mm}
\resizebox{1.0\linewidth}{!}{
\begin{tabular}{cc|ccc}
\toprule
    $\textbf{Dataset}$    &     $\textbf{Model}$     & $\textbf{Baseline}$     & $\textbf{w/ GNN-SD}$     &     $\textbf{Gain}$     \\
% \hline
\midrule
% \midrule
\multirow{5}{*}{ENZYMES}        & GCN               & $64.00_{\pm5.63}$     & $66.66_{\pm3.94}$       &$(+2.66)$            \\
                                & GAT               & $65.33_{\pm5.90}$     & $68.00_{\pm2.66}$       &$(+2.66)$                                         \\
                                & GraphSage         & $68.33_{\pm6.41}$     & $\mathbf{70.00}_{\pm5.05}$       &$(+1.66)$          \\
                                & GIN               & $66.00_{\pm7.19}$     & $\textbf{69.33}_{\pm4.02}$       &$(+3.33)$                 \\
                                & GatedGCN          & $65.33_{\pm4.52}$     & $67.33_{\pm1.33}$       &$(+2.00)$                                         \\
\midrule
\multirow{5}{*}{DD}             & GCN               & $77.83_{\pm1.02}$     & $\mathbf{78.67}_{\pm1.68}$       &$(+0.84)$  \\
                                & GAT               & $76.65_{\pm2.51}$     & $77.50_{\pm2.50}$       &$(+0.85)$                               \\
                                & GraphSage         & $76.14_{\pm1.66}$     & $77.49_{\pm1.89}$       &$(+1.35)$                              \\
                                & GIN               & $73.00_{\pm3.90}$     & $74.77_{\pm3.50}$       &$(+1.77)$                               \\
                                & GatedGCN          & $77.83_{\pm1.67}$     & $\textbf{78.20}_{\pm{1.98}}$       &$(+0.37)$            \\
\midrule
\multirow{5}{*}{PROTEINS}       & GCN               & $75.55_{\pm2.91}$     & $76.81_{\pm3.19}$       &$(+1.26)$  \\
                                & GAT               & $76.36_{\pm2.77}$     & $\mathbf{77.53}_{\pm3.38}$       &$(+1.17)$    \\
                                & GraphSage         & $75.55_{\pm4.02}$     & $76.71_{\pm3.81}$       &$(+1.15)$                           \\
                                & GIN               & $64.86_{\pm3.03}$     & $70.06_{\pm4.89}$       &$(+5.20)$                           \\
                                & GatedGCN          & $76.36_{\pm3.94}$     & $\textbf{76.90}_{\pm3.68}$       &$(+0.54)$   \\
\bottomrule
\end{tabular}
} 
% \vspace{-2mm}
\label{tab:table1-mainComparison}%
\vspace{-1mm}
\end{table}

\begin{table}[thbp] %[hth]
\centering
\caption{Compatibility study. Experiments are conducted under  the full-supervised scheme, following DropEdge's implementation. 
}
\vspace{-2mm}
% \vspace{0.5mm}
\resizebox{0.93\linewidth}{!}{
\begin{tabular}{c|cccc}
\toprule
    \multirow{2}{*}{$\textbf{Models}$}  &         \multicolumn{4}{c}{$\textbf{Layers}$}  \\
                            &   8 &    16   &   32   &   50    \\
\midrule
   Baseline   &   78.1$_{\pm0.9}$    &     79.1$_{\pm1.1}$        &       79.3$_{\pm0.8}$        &   78.4$_{\pm1.1}$    \\
% \midrule
    w/ DropEdge &  79.4$_{\pm0.7}$    &    79.3$_{\pm0.9}$      &       79.4$_{\pm1.2}$      &   78.7$_{\pm0.9}$    \\
    w/ GNN-SD   &  79.1$_{\pm0.8}$    &    \textbf{79.9}$_{\pm0.5}$      &       79.7$_{\pm1.1}$        &   79.2$_{\pm0.8}$    \\
    w/ Both     &  \textbf{80.1}$_{\pm0.6}$    &  79.6$_{\pm0.8}$        &       \textbf{80.2}$_{\pm0.8}$        &   \textbf{79.6}$_{\pm0.5}$    \\
% \hline
\bottomrule
\end{tabular}
}
\label{tab:with-dropedge}
\vspace{-3mm}
\end{table}

\paragraph{Compatibility Evaluation}
There exist other methods that aim at facilitating the training of GNNs. One influential work is DropEdge~\cite{rong2019dropedge}, which randomly samples graph edges to introduce data augmentation.
We conduct experiments on JKNet~\cite{xu2018representation} and Citeseer dataset to evaluate the compatibility of these orthogonal training schemes.
The edge sampling ratio in DropEdge is searched at the range of $\{0.1,...,0.9\}$, with results demonstrated in Table~\ref{tab:with-dropedge}. 
It reads that while both GNN-SD and DropEdge are capable of improving the training, GNN-SD might perform better on deep backbones. Notably, employing them concurrently is likely to  deliver further promising enhancement.

\begin{table}[ht] %[hth]
\centering
% \vspace{0.5mm}
\caption{Ablation study of  different knowledge sources. 
        }
\vspace{-2mm}
\resizebox{0.97\linewidth}{!}{
\begin{tabular}{c|c|ccc|cc}
\toprule
                        &   $\textbf{GNN}$-$\textit{\textbf{B}}$ & $\textbf{GNN}$-$\textit{\textbf{L}}$ & $\textbf{GNN}$-$\textit{\textbf{N}}$ & $\textbf{GNN}$-$\textit{\textbf{G}}$  &  \multicolumn{2}{c}{$\textbf{GNN}$-$\textit{\textbf{M}}$}    \\ % & Best Model\\
\midrule
    $\mathcal{L}_{\mathrm{L}}$ &    & $\surd$  &   &   & $\surd$  & $\surd$ \\% & \\
    $\mathcal{L}_{\mathrm{N}}$  &    &   & $\surd$  &   & $\surd$  & $\surd$ \\%&  \\
    $\mathcal{L}_{\mathrm{G}}$  &    &   &   & $\surd$  &    & $\surd$    \\%&  \\
    % $\textit{L}_\text{total}$  &   &   &   &   &   $\surd$   &   \\
% \hline
% \hline
\midrule
        Cora     & $80.12$   & $\cellcolor{lightgray}{79.83}$  & $80.83$  & /  &  $\textbf{\cellcolor{rgb:red!10,0.1;green!10,0.45;blue!10,0.85}{81.16}}$     & / \\  
        Citeseer & $70.53$   & $71.16$  & $71.43$  &  /  & $\textbf{\cellcolor{rgb:red!10,0.1;green!10,0.45;blue!10,0.85}{71.52}}$    & / \\
        Pubmed    & $78.52$   & $\cellcolor{lightgray}{78.53}$  & $\textbf{79.44}$  & /    & $79.26$    & / \\
\midrule
        ENZYMES     & $66.00$   & $66.66$  & $67.66$  & $67.66$  &  $66.66$    & $\textbf{\cellcolor{rgb:red!10,0.1;green!10,0.45;blue!10,0.85}{69.33}}$  \\ 
        DD          & $73.00$   & $73.76$  & $\textbf{74.77}$  &  $73.51$  &   $73.76$  &  $74.14$ \\
\bottomrule
\end{tabular}
}
\label{tab:table2-differentSettings}
\vspace{-3mm}
\end{table}
\subsection{Ablation Studies}
We perform an ablation study to evaluate the knowledge sources and identify the effectiveness of our core technique, as shown in Table~\ref{tab:table2-differentSettings}.
We select GAT as the evaluation backbone for node classification and GIN for graph classification.
We name the baseline as `GNN-$\textit{B}$', and model solely distilled by intermediate logits~\cite{zhang2019your}, neighborhood discrepancy, and compact graph embedding as `GNN-$\textit{L}$', `GNN-$\textit{N}$', `GNN-$\textit{G}$', respectively.
The models distilled by mixed knowledge are represented as `GNN-$\textit{M}$'.
One observation from the results is that simply adopting the intermediate logits seems to fail in bring consistent improvement (highlighted in gray),  while it may cooperate well with other sources since it promotes the updating of shallow features (highlighted in blue). 
In contrast, the  discrepancy retaining plays the most important role in distillation training.
For graph classifications, the involvement of compact graph embedding also contributes well while jointly works with the others.

\section{Conclusion}\label{sec:conclusion} %%\vspace{-1mm}
We have presented an efficient yet generic  GNN Self-Distillation (GNN-SD) framework tailored for boosting GNN performance. 
Experiments verify that it achieves state-of-the-art distillation performance.
Meanwhile, serving as a drop-in replacement of the standard training process, it yields consistent and considerable enhancement on various GNN models.

\bibliographystyle{named}
\bibliography{ijcai21}

\clearpage
\appendix
    \onecolumn
    \begin{center}
    \huge
    \textbf{Appendix}
     \\[18pt]
    \end{center}
% ]

\section{Proof of Proposition 1}\label{proof-1}
For simplicity, here we use the $L_p$-norm to measure the vector difference. Pair-wise distance metric $d_2(\BX, \mathcal{G})$ calculates the total sum of all node pairs dissimilarity, as follows:
\begin{align}
    d_2(\BX, \mathcal{G}) &=    \notag \frac{1}{|\mathcal{E}|}\sum_{v\in\mathcal{V}}\sum_{c\in\mathcal{N}(v)}\vert\vert \BX_{v\cdot}-\BX_{c\cdot}\vert\vert_p \\ \notag
    & \geq \frac{1}{|\mathcal{E}|}\sum_{v\in\mathcal{V}}\vert\vert \sum_{c\in\mathcal{N}(v)}(\BX_{v\cdot}-\BX_{c\cdot})\vert\vert_p \\    \notag
    &= \frac{1}{|\mathcal{E}|}\sum_{v\in\mathcal{V}}\vert\vert \BD_{v,v}(\BX_{v\cdot}-\frac{(\BA\BX)_{v\cdot}}{\BD_{v,v}})\vert\vert_p \\  
    & \geq \frac{1}{|\mathcal{E}|}\sum_{v\in\mathcal{V}}\vert\vert \BX_{v\cdot} - (\BD^{-1}\BA\BX)_{v\cdot} \vert\vert_p
\end{align}
The RHS of the last inequality computes the sum of dissimilarity between nodes and their 1-hop aggregated neighbors, formulating the neighbor-wise distance $d_1(\BX, \mathcal{G})$. The proof is concluded.

\section{Proof of Proposition 2}\label{proof-2}
The mutual information between the NDR of consecutive layers is defined as:
\begin{align}
\label{mutual-information}
    I(\mathbf{s}^\pare{l}, \mathbf{s}^\pare{l+1}) &= H(\mathbf{s}^\pare{l}) - H(\mathbf{s}^\pare{l}|\mathbf{s}^\pare{l+1})  \notag \\ 
    &= -\mathbbm{E}_{\mathbf{s}^\pare{l}}[\mathrm{log}p(\mathbf{s}^\pare{l})] + \mathbbm{E}_{\mathbf{s}^\pare{l},\mathbf{s}^\pare{l+1}}[\mathrm{log}p(\mathbf{s}^\pare{l}|\mathbf{s}^\pare{l+1})],
\end{align}
since the true conditional probability is not intractable, we resort to the help with the variational distribution of $q(\mathbf{s}^\pare{l}|\mathbf{s}^\pare{l+1})$ that approximates $p(\mathbf{s}^\pare{l}|\mathbf{s}^\pare{l+1})$, and study the lower bound of the mutual information.
Continuing from formula (\ref{mutual-information}),
\begin{align}
\label{variational-approximation}
    I(\mathbf{s}^\pare{l}, \mathbf{s}^\pare{l+1}) &= H(\mathbf{s}^\pare{l})  + \mathbbm{E}_{\mathbf{s}^\pare{l},\mathbf{s}^\pare{l+1}}[\mathrm{log}p(\mathbf{s}^\pare{l}|\mathbf{s}^\pare{l+1})] \notag \\
    &= H(\mathbf{s}^\pare{l}) + \mathbbm{E}_{\mathbf{s}^\pare{l},\mathbf{s}^\pare{l+1}}[\mathrm{log}q(\mathbf{s}^\pare{l}|\mathbf{s}^\pare{l+1})] + \mathbbm{E}_{\mathbf{s}^\pare{l}}[\mathrm{D_{KL}}(p(\mathbf{s}^\pare{l}|\mathbf{s}^\pare{l+1})||q(\mathbf{s}^\pare{l}|\mathbf{s}^\pare{l+1}))] \notag \\
    &\geq H(\mathbf{s}^\pare{l}) +  \mathbbm{E}_{\mathbf{s}^\pare{l},\mathbf{s}^\pare{l+1}}[\mathrm{log}q(\mathbf{s}^\pare{l}|\mathbf{s}^\pare{l+1})],
\end{align}
here we can adopt the Gaussian distribution with  heteroscedastic expectation $\mu=\mathbf{s}^\pare{l+1}$ and variance $\sigma={\mathbf{1}_{N}}^{\text{T}}\BD$ as the variational distribution. 
And it's assumed with the property that NDR of nodes are conditionally independent given the ones of subsequent layer. 
That is, $q(\mathbf{s}^\pare{l}|\mathbf{s}^\pare{l+1})=\prod_{v}q(S_v^\pare{l})|S_v^\pare{l+1})=\prod_v\frac{1}{\BD_{v,v}\sqrt{2\pi}}\mathrm{exp}(-\frac{(S_v^\pare{l}-S_v^\pare{l+1})^2}{2\BD_{v,v}^2})$.
Then, the logarithm of variational distribution is expressed as:
\begin{align}
\label{gaussian-variational}
    \mathrm{log}q(\mathbf{s}^\pare{l}|\mathbf{s}^\pare{l+1})
    &= \sum_{v=1,...,N}\mathrm{log}q(S^\pare{l}_v|S_v^\pare{l+1}) \notag \\
    &= \sum_{v=1,...,N}-\mathrm{log}\frac{1}{\BD_{v,v}} - \frac{(S^\pare{l}_v-S^\pare{l+1}_v)^2}{1/\BD_{v,v}} +\text{constant}.
\end{align}
Since the node degree $\BD_{v,v}$ is fixed and $-\mathrm{log}\frac{1}{\BD_{v,v}}$ is greater than zero, then combining equation (\ref{variational-approximation}) and (\ref{gaussian-variational}) leads to:
\begin{equation}
    I(\mathbf{s}^\pare{l}, \mathbf{s}^{\pare{l+1}})\geq H(\mathbf{s}^\pare{l}) -  \mathbbm{E}_{\mathbf{s}^\pare{l},\mathbf{s}^\pare{l+1}}[\Vert\BD(\mathbf{s}^\pare{l+1}-\mathbf{s}^\pare{l})^{\text{T}}\Vert_2^2].
\end{equation}
The proof is concluded.

\section{Experimental Environments}
Most experiments conducted in this paper are run on a NVIDIA 2080 Ti GPU with 11 GB memory, except for very-deep GNNs, which are conducted on a NVIDIA Tesla P40 with 24GB memory.
Experiments are mainly implemented by PyTorch of version 1.3.1 and  DGL~\cite{wang2019dgl} of version 0.4.2.

\section{Comparison on Two Supervision Schemes of Retaining NDR}\label{append:two-scheme}
Table \ref{tab:two-scheme} compares the results of two supervision schemes: 
Transferring the NDR of shallow layers to deep ones and the reverse strategy.
Since the rationale of our design is to retain high neighborhood discrepancy from shallow layers to deep layers, 
the common strategy for CNN models that transfers knowledge from deep layers to shallow layers is not the desired scheme for conducting GNN-SD. 
One can see that performing neighborhood discrepancy retaining by serving the NDR of deep embeddings as supervision targets is not optimal, sometimes even detrimental for GNN models.
These results verify  our conjecture. 

\begin{table}[h] %[hth]
\centering
\caption{Comparison of two schemes of NDR transferring, in accuracy gain ($\%$). Performing neighborhood discrepancy retaining by serving the NDR of embeddings in the shallow layer as supervision signals is denoted as `shallow2deep', and the other one is `deep2shallow'. A 3-layer GAT is selected as the baseline.
}
\vspace{-2mm}
% \vspace{0.5mm}
\resizebox{0.35\linewidth}{!}{
\begin{tabular}{c|cccc}
\toprule
    GAT  &         Cora     &   Citeseer    & Pubmed    \\
\midrule
   shallow2deep   & +0.66    &   +0.73    &  +0.80      \\
% \midrule
   deep2shallow &  -0.52    &    -0.22    &  +0.17        \\
% \hline
\bottomrule
\end{tabular}
}
\label{tab:two-scheme}
\vspace{-2mm}
\end{table}

\section{The Generic GNN Self-Distillation Framework}\label{append:framework}
Recall the summarized  self-distillation objective,
\begin{equation}\notag
\mathcal{M}_{\mathrm{SD}}^{g, L}: \mathcal{T}_{s}(\mathcal{C}_{s}(\mathcal{G}, P_s))\to \mathcal{T}_{t}(\mathcal{C}_{t}(\mathcal{G}, P_t)),
\end{equation}
the various combination of  granularity settings and transformation functions leave wide space for further exploration.
Generally, the choices of granularities of a graph $\mathcal{C}(\mathcal{G}, P)$ constitute the set of
$\{\BX^\pare{P}, \BE^\pare{P}, \mathbf{SG}^\pare{P}, \BG^\pare{P}\}$, which is made up of: 
1) fine-grained embeddings at node and edge levels, denoted as $\BX$ and $\BE$, 
2) coarse-grained embeddings at full-graph and sub-graph levels, denoted as  $\BG\in {\R}^{1\times d}$  and $\mathbf{SG}\in \mathbb{R}^{s\times d}$ of:
\begin{align}\label{eq:graph-readout}
    & \BG = \mathrm{Readout}_{v\in \mathcal{G}}(\BX_{v\cdot}) \notag \\
    & \mathbf{SG}_{i\cdot}=\mathrm{Readout}_{v\in \mathcal{SG}_i}(\BX_{v\cdot}), \quad i=1,2,...,s
\end{align}
where $s$ represents the number of partitioned sub-graphs, $\mathcal{SG}_i=\{\mathcal{V}_{i}, \mathcal{E}_{i}, \BA_{i}\}$ is the $i$th sub-graph  that consists of the corresponding vertex and edge subsets and adjacency.
The straightforward approach to generate sub-graph is to perform random sampling schemes on the original graph. 
It remains an interesting problem that how to sample representative nodes in cooperate with the distillation target.

In practice, the chosen granularity depends on the specific scenario and goal that wishes to obtain by self-distillation.
The transformation should comprise between the properties of `easy to learn' and `avoid information missing'.
In the manuscript, we have adopted the fine-grained node embeddings $\BX$ and coarse-grained full-graph embedding $\BG$ to perform a mixed knowledge self-distillation for training GNNs.
The core technique of discrepancy retaining performs distillation in a progressive manner, results in the  objective instance of:
\begin{equation}\label{eq:Mapping-node}
\mathcal{M}_{\mathrm{node}}^{g, L}: \mathrm{NDR}(\BX^\pare{l+1})\to \mathrm{NDR}(\BX^\pare{l}),
\end{equation}
where $\mathrm{NDR}(\cdot)$ is formulated by Eq.(\ref{eq:node-smooth-i}) and Eq.(\ref{eq:node-smooth}).
We have further involved knowledge sources of the full-graph embedding in our GNN-SD framework, as illustrated in formula~(\ref{eq:Mapping-graph}).

In fact, the training objective of formula (\ref{eq:Mapping-graph}) can be naturally extended to sub-graph level by replacing $\BG$ by $\mathbf{SG}$. 
For applying the GNN-SD to edge features $\BE$, a simple two-step solution is to firstly transform the node-adjacency matrix $\BA$ into edge-adjacency matrix of $\BA^e$, where each element is:
\begin{equation}\label{eq:edge-adjacency}
    [\BA^e]_{i,j}= \begin{cases}
                1& i\cap j\neq \emptyset,\;\text{edge $i$, $j$ share a same node} \\
                0& \text{otherwise}                             
                \end{cases}         
\end{equation}
and then extend the NDR to edge-level by replacing Eq.(\ref{eq:node-smooth-i}) with:
\begin{equation}\label{eq:edge-smooth-i}
S_i^e = 1 - \frac{{\BE_{i\cdot}}^\top{(\BA^e\BE)}_{i\cdot}}{||\BE_{i\cdot}||_2\cdot||(\BA^e\BE)_{i\cdot}||_2}, i=1,...,M.
\end{equation}

It's worthy noting that the  transformation $\mathcal{T}$ can not only be manually designed but  learned during the training process.
 A linear layer or $\mathrm{MLP}$ is able to  facilitate the student layers to match the target knowledge, which leads  formula (\ref{eq:Mapping-node}), for example,  to the following variant:
 \begin{equation}\label{eq:Mapping-node-learn}
\mathcal{M}_{\mathrm{node}}^{g, L}: \mathrm{NDR}(f(\BX^\pare{P_s}))\to \mathrm{NDR}(\BX^\pare{P_t}), 
\end{equation}
where $f$ denotes the learnable function. We empirically find that this technique is typically useful for distillation on large-scale datasets and very-deep GNNs. 

\section{Additional NDR Curves Exemplars}\label{append:explamar}
\begin{figure*}[h]
  \centering
  % Requires \usepackage{graphicx}
  \includegraphics[width=0.65\linewidth]{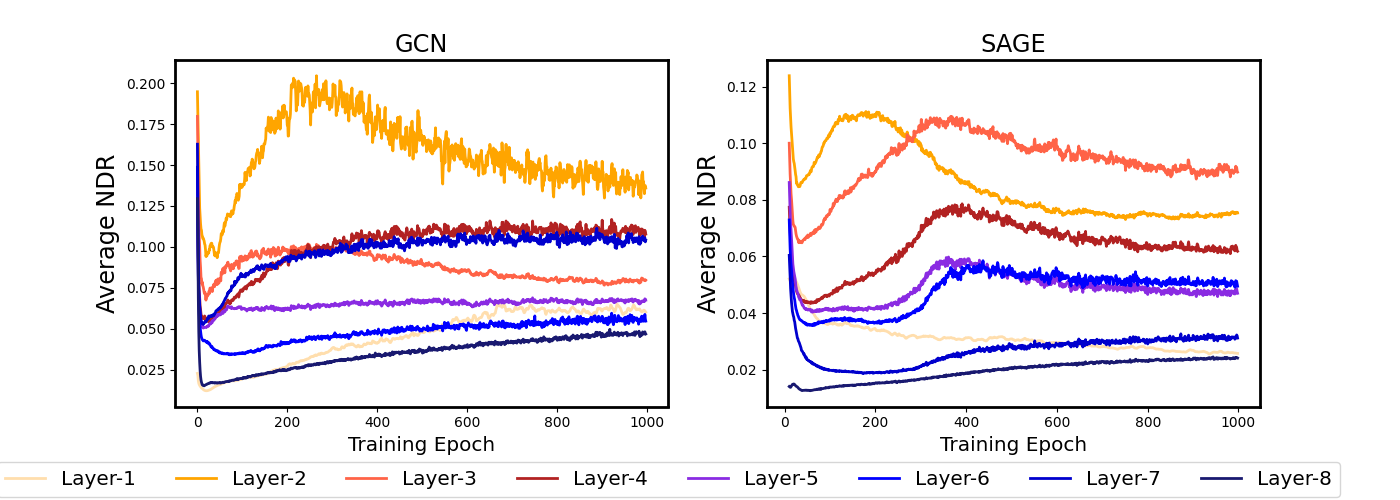}%\vspace{-2mm}
%   \vspace{-2mm}
  \caption{NDR curves of the two 8-layers GNNs. Smaller value meanssuffering from more over-smoothing.}
 \label{append:fig} 
 \vspace{-2mm}
\end{figure*}

Figure~\ref{append:fig} shows the NDR Curves of the 8-layer GCN and GraphSage on OGB-arxiv dataset. It's observed that the initial embedded graph (in light yellow) has low NDR value (in average), which might be caused by the fact that the first GNN layer takes much more noise.   
In GCN, the embedded graph of the  second layer (in orange) possesses the highest NDR across the training stage, while in GraphSage it appears alternation that the third layers (in tomato) become more discriminated in the late training stage. 
The above observations motivate us to adaptively select the initial target supervision signal.

Furthermore, we find that the consecutive intermediate layers could show minor overlap in the average NDR, and it doesn't make sense to perform knowledge retaining for the case that the target layer show lower neighborhood discrepancy than the online layer. Thus we introduce explicit  teacher selection in the ADR regularizer.
% \vspace{-3mm}

We provide a comparison study to show the effect of the adaptive distillation strategy, with results in Table \ref{tab:comp-ada-naive}. Here, the `naive matching' denotes that the distillation loss is defined as $\mathcal{L}_{\mathrm{N}} = \sum_{l=1,...,L-1}d^2(\mathbf{s}^\pare{l+1}, \mathbf{s}^\pare{l})$.
% \vspace{-2mm}

\begin{table}[h] %[hth]
\centering
\caption{The effect of the adaptive self-distillation.}
\vspace{-2mm}
% \vspace{0.5mm}
\resizebox{0.76\linewidth}{!}{
\begin{tabular}{c|cc|c|cc}
\toprule
    OGB-arxiv  &         GCN     &   GraphSage     & Cora   & GCN   & GraphSage\\
\midrule
    baseline    & 71.84$_{\pm0.18}$  & 71.12$_{\pm0.38}$    &   baseline    &  80.67$_{\pm1.14}$    &      79.28$_{\pm1.91}$   \\ 
   naive matching    &  72.08$_{\pm0.20}$    &  71.62$_{\pm0.32}$   & naive matching  & 82.70$_{\pm0.97}$   &   79.97$_{\pm1.81}$  \\
% \midrule
   adaptive matching &  72.34$_{\pm0.32}$     &   71.88$_{\pm0.23}$  & adaptive matching &  82.92$_{\pm0.89}$    &  80.31$_{\pm2.35}$        \\
% \hline
\bottomrule
\end{tabular}
}
\label{tab:comp-ada-naive}
\vspace{-2mm}
\end{table}

\section{Details of the Comparison with Other Distillation Methods}\label{append:LSP}
For processing the PPI dataset, where possesss relative large scale graphs, we adopt a learnable linear layer in the knowledge transformation function to facilitate knowledge transfer between layers, and thus the resulting distillation object is the same as formula (\ref{eq:Mapping-node-learn}).

Table~\ref{tab:referLSP} describe the hyper-parameters of the teacher model used in those teacher-student KD methods, and the ones of baseline student model.
One can notice that the teacher model is set in an fatter architecture with fewer layers compared with the student, which is not as usual in the conventional settings on CNNs. This confirms our  concern about  the difficulty of selecting a qualified teacher GNN model.
In Table~\ref{tab:supp-compCVPR}, we provide the detailed four runs results of the re-implemented  baseline backbone used in LSP and our GNN-SD method.
% \vspace{-2mm}

\begin{table}[h]
\centering
\caption{Settings of the baseline backbone used on the PPI dataset for comparing with LSP.}
\vspace{-2mm}
\resizebox{0.535\linewidth}{!}{
\begin{tabular}{c|cccc}
\toprule
        Model    &   Layers & Attention heads & Hidden features & \#Params \\
\midrule
    Teacher ($T$)   & 3 & 4,4,6 &   256,256,121 & 3.64M \\
    Baseline   &  5 & 2,2,2,2,2    &   68,68,68,68,121 & 0.16M   \\
\bottomrule
\end{tabular}
}
\label{tab:referLSP}
\vspace{-2mm}
\end{table}
% \vspace{-2mm}

\begin{table}[h] %[hth]
\centering
\caption{Performance comparison on the PPI dataset. The second column records the experiment results reported in the paper of LSP. In the other columns, we report the overall running results of 4 runs of our experiments. }
\vspace{-2mm}
\resizebox{0.48\linewidth}{!}{
\begin{tabular}{c|c|ccccc}
\toprule
        \multirow{4}{*}{Method}    &   \multicolumn{6}{c}{F1 Score}    \\
                                    \cmidrule{2-3} \cmidrule{4-7}       
                                    \cmidrule{2-3} \cmidrule{4-7}   
                                 &   \multirow{2}{*}{Report}   &   \multicolumn{5}{c}{our runnning results} \\
                                    \cmidrule{3-7}
                                    \cmidrule{3-7}
                                      &     &   Avg.     & (1st  & 2nd  & 3rd  & 4th )         \\
\midrule
    Baseline  &   $95.7^\star$                 &       \multicolumn{1}{c}{95.6} & (95.64 &  95.83   &   95.27   &   95.69)  \\
    GNN-SD         &   /                           &       \multicolumn{1}{c}{$\textbf{96.2}$} &   (96.14   &   96.24   &   96.22   &   96.20)   \\
% \hline
\bottomrule
\end{tabular}       
}
\label{tab:supp-compCVPR}%\vspace{-3mm}
\vspace{-2mm}
\end{table}

% \clearpage
\section{Hyper-parameter Settings of the Node Classification Experiments, Table \ref{tab:various-depth}}\label{append:node-hyper}
Table \ref{tab:notation} gives the description of the meaning of the hyper-parameters.
We search the hyper-parameters for GNNs under varying depths to obtain the fair and strong baselines. 
For generating intermediate logits, we leverage a sharing-weights 2-layer $\mathrm{MLP}$ to take intermediate graph embeddings as input. 
Loss weight of $\alpha$ is searched at the range of $\{0, 0.001, 0.01, 0.1, 1\}$, $\beta$ in $\{0.001, 0.01, 0.1, 1, 10, 100\}$.
Table \ref{tab:hypers2} summarizes the detail settings.

\begin{table}[h!]
\centering
\caption{Hyper-parameter Description}
% \vspace{-1mm}
\resizebox{0.37\linewidth}{!}{
\begin{tabular}{cc}
\toprule
        Hyperparameter    &   Description  \\
\midrule
    lr          &   learning rate         \\
    weight-decay &  L2 regularization weight            \\
    dropout &       dropout rate       \\
    % withbn &        using batch normalization      \\
    % withloop    &   adding self-feature to formulate a node loop    \\
    \#layer &   number of hidden layers         \\
    \#hidden &    hidden dimensionality        \\
    \#head   &   number of attention heads        \\
    \#epoch & number of training epochs   \\
    $\alpha$, $\beta$, $\gamma$  & loss weight in Eq.(\ref{eq:total-object}) \\
\bottomrule
\end{tabular}
}
\label{tab:notation}%\vspace{-3mm}
\end{table}
\begin{table*}[h] %[hth]
\centering
\caption{The hyper-parameters of each backbones under varying depths for citation datasets classificatoin, and the selected optimal loss weights.}
\vspace{-2mm}
\resizebox{1.0\linewidth}{!}{
\begin{tabular}{c|c|c|c|c}
\toprule
        Dataset    &   Backbone   & Layers  &    Model Hyperparameters   & Loss Hyperparameters\\   % &    Acc.$\pm$s.d. 
\midrule
        \multirow{11}{*}{Cora}    &   \multirow{4}{*}{GAT}   &   2    & lr=1e-2, weight-decay=1e-3, dropout=0.5, \#hidden=64, \#head=4, \#epoch=300 &  $\alpha=0.01, \beta=0.001$ \\
        \cmidrule{3-5}
                                &                       &   4    & lr=1e-2, weight-decay=1e-2, dropout=0.5, \#hidden=128, \#head=4, \#epoch=300 &  $\alpha=1.0, \beta=1.0$  \\
        \cmidrule{3-5}
                                &                       &   8        & lr=1e-2, weight-decay=1e-3, dropout=0.1,   \#hidden=64, \#head=4, \#epoch=300 &   $\alpha=1.0, \beta=1.0$  \\
        \cmidrule{3-5}
                                &                       &   16       &  lr=1e-2, weight-decay=1e-2, dropout=0.1,   \#hidden=16, \#head=4, \#epoch=300 &   $\alpha=0.001, \beta=0.1$  \\
        \cmidrule{2-5}
                                &   \multirow{4}{*}{GraphSage}   &   2   & lr=1e-2, weight-decay=1e-3, dropout=0.5,   \#hidden=64,  \#epoch=300 &  $\alpha=0, \beta=0.001$ \\
        \cmidrule{3-5}
                                &                       &   4         & lr=1e-2, weight-decay=1e-3, dropout=0.5,   \#hidden=64, \#head=4, \#epoch=300 &  $\alpha=0.01, \beta=1.0$  \\
        \cmidrule{3-5}
                                &                       &   8         & lr=1e-2, weight-decay=1e-3, dropout=0.1,   \#hidden=64, \#epoch=300 &   $\alpha=1.0, \beta=0.001$  \\
        \cmidrule{3-5}
                                &                       &   16        &  lr=1e-3, weight-decay=1e-2, dropout=0.1,   \#hidden=128, \#epoch=300 &   $\alpha=0.001, \beta=0.001$  \\
\midrule
        \multirow{11}{*}{Citeseer}    &   \multirow{4}{*}{GAT}   &   2    & lr=1e-2, weight-decay=1e-3, dropout=0.8,   \#hidden=128, \#head=4, \#epoch=300 &  $\alpha=0, \beta=0.001$ \\
        \cmidrule{3-5}
                                &                       &   4    & lr=1e-2, weight-decay=1e-2, dropout=0.6,   \#hidden=256, \#head=4, \#epoch=300 &  $\alpha=1.0, \beta=0.01$  \\
        \cmidrule{3-5}
                                &                       &   8        & lr=1e-2, weight-decay=1e-2, dropout=0.1,   \#hidden=128, \#head=4, \#epoch=300 &   $\alpha=0.1, \beta=0.01$  \\
        \cmidrule{3-5}
                                &                       &   16       &  lr=1e-2, weight-decay=1e-4, dropout=0.1,   \#hidden=16, \#head=4, \#epoch=300 &   $\alpha=0, \beta=1.0$  \\
        \cmidrule{2-5}
                                &   \multirow{4}{*}{GraphSage}   &   2   & lr=1e-2, weight-decay=1e-2, dropout=0.6,   \#hidden=128, \#epoch=300 &  $\alpha=0, \beta=10.0$ \\
        \cmidrule{3-5}
                                &                       &   4         & lr=1e-2, weight-decay=1e-3, dropout=0.2,   \#hidden=128, \#head=4, \#epoch=300 &  $\alpha=0, \beta=0.01$  \\
        \cmidrule{3-5}
                                &                       &   8         & lr=1e-2, weight-decay=1e-3, dropout=0.1,   \#hidden=128, \#epoch=300 &   $\alpha=0.001, \beta=0.1$  \\
        \cmidrule{3-5}
                                &                       &   16        &  lr=1e-3, weight-decay=1e-2, dropout=0.1,   \#hidden=128, \#epoch=300 &   $\alpha=0.01, \beta=0.001$  \\
\midrule
        \multirow{11}{*}{Pubmed}    &   \multirow{4}{*}{GAT}   &   2    & lr=1e-2, weight-decay=1e-3, dropout=0.1,   \#hidden=64, \#head=4, \#epoch=300 &  $\alpha=0.1, \beta=0.1$ \\
        \cmidrule{3-5}
                                &                       &   4    & lr=1e-2, weight-decay=1e-2, dropout=0.1,   \#hidden=128, \#head=4, \#head=4, \#epoch=300 &  $\alpha=0, \beta=10.0$  \\
        \cmidrule{3-5}
                                &                       &   8        & lr=1e-2, weight-decay=1e-2, dropout=0.1,   \#hidden=64, \#head=4, \#epoch=300 &   $\alpha=0, \beta=0.1$  \\
        \cmidrule{3-5}
                                &                       &   16       &  lr=1e-3, weight-decay=1e-3, dropout=0.1,   \#hidden=128, \#head=4, \#epoch=300 &   $\alpha=1.0, \beta=100.0$  \\
        \cmidrule{2-5}
                                &   \multirow{4}{*}{GraphSage}   &   2   & lr=1e-2, weight-decay=1e-3, dropout=0.2,   \#hidden=128, \#epoch=300 &  $\alpha=0.01, \beta=0.01$ \\
        \cmidrule{3-5}
                                &                       &   4         & lr=1e-2, weight-decay=1e-2, dropout=0.1,   \#hidden=64, \#head=4, \#epoch=300 &  $\alpha=0.1, \beta=0.001$  \\
        \cmidrule{3-5}
                                &                       &   8         & lr=1e-3, weight-decay=1e-2, dropout=0.3,   \#hidden=64, \#epoch=300 &   $\alpha=0.1, \beta=10.0$  \\
        \cmidrule{3-5}
                                &                       &   16        &  lr=1e-3, weight-decay=1e-2, dropout=0.1,   \#hidden=128, \#epoch=300 &   $\alpha=1.0, \beta=100.0$  \\
\bottomrule
\end{tabular}
}
\label{tab:hypers2}
\vspace{-3mm}
\end{table*}

\clearpage
\section{Hyper-parameter Settings of the Graph Classification Experiments, Table \ref{tab:table1-mainComparison}}\label{append:graph-hyper}
We adopt the Adam optimizer~\cite{kingma2014adam} for model training in this paper. 
For conducting 10-fold cross validation on graph kernel datasets, the random seed is fixed for reproducing the results.
The hyper-parameters (e.g. the number of hidden dimensions) of the baseline models are set for matching the budget that each model contains around $100$k parameters.
The $\mathrm{Readout}$ operator in Eq.(\ref{ref:graph_readout}) is set as the same as the one used in the baseline GNN backbone.
For generating intermediate logits, we leverage a sharing-weights 2-layer $\mathrm{MLP}$ to take intermediate graph embeddings as input. Specially, GIN model which has already contained intermediate prediction layers to generate residual probability scores for the final output distribution, no extra $\mathrm{MLP}$ is introduced.
For the  choice of the hyper-parameters of $\alpha$, $\beta$ and $\gamma$ in Eq.(\ref{eq:total-object}), we carried out a simple grid search at the range  of $\alpha\in\{0, 0.1, 1\}$, $\beta\in\{0, 0.1, 1\}$, $\gamma\in\{0, 0.01, 0.1, 1\}$ for each controlled experiments.
Then, the detailed hyper-parameters are summarized in Table \ref{tab:hypers}.

\begin{table*}[htbp] %[hth]
\centering
\caption{The hyper-parameters of each backbones on graph kernel datasets  for matching the baseline budgets, and the selected optimal loss weights.}
% \vspace{0.5mm}
\resizebox{1.0\linewidth}{!}{
\begin{tabular}{c|c|c|c}
\toprule
        Dataset    &   Backbone   &  Model Hyperparameters   & Loss Hyperparameters\\   % &    Acc.$\pm$s.d. 
\midrule
        \multirow{5}{*}{ENZYMES}    &   GCN      & lr=7e-4, weight-decay=0, dropout=0, \#layer=4, \#hidden=128, \#epoch=1000 &  $\alpha=0, \beta=0.1, \gamma=0$ \\
        \cmidrule{2-4}
                                &   GAT     & lr=1e-3, weight-decay=0, dropout=0, \#layer=4, \#hidden=16, \#head=8, \#epoch=1000 &  $\alpha=1, \beta=0.1, \gamma=0$  \\
        \cmidrule{2-4}
                                &   GraphSage     & lr=7e-4, weight-decay=0, dropout=0, \#layer=4, \#hidden=96, \#epoch=1000 &   $\alpha=0, \beta=1, \gamma=0$  \\
        \cmidrule{2-4}
                                &   GIN    &  lr=7e-3, weight-decay=0, dropout=0, \#layer=4, \#hidden=96, \#epoch=1000 &   $\alpha=0.1, \beta=1, \gamma=1$  \\
        \cmidrule{2-4}
                                &   GatedGCN    &  lr=7e-4, weight-decay=0, dropout=0, \#layer=4, \#hidden=64, \#epoch=1000 &   $\alpha=1, \beta=0.1, \gamma=0$   \\
\midrule
        \multirow{5}{*}{DD}    &   GCN   &  lr=1e-5, weight-decay=0, dropout=0, \#layer=4, \#hidden=128, \#epoch=1000  &  $\alpha=0, \beta=0, \gamma=0.01$  \\
        \cmidrule{2-4}
                                    &   GAT     &  lr=5e-5, weight-decay=0, dropout=0, \#layer=4, \#hidden=16,  \#head=8, \#epoch=1000 & $\alpha=0, \beta=0.1, \gamma=0$  \\ 
        \cmidrule{2-4}
                                    &   GraphSage    & lr=1e-5, weight-decay=0, dropout=0, \#layer=4, \#hidden=96, \#epoch=1000 & $\alpha=1, \beta=0, \gamma=0$  \\ 
        \cmidrule{2-4}
                                    &   GIN    &  lr=1e-3, weight-decay=0, dropout=0, \#layer=4, \#hidden=96, \#epoch=1000 & $\alpha=0, \beta=1, \gamma=0$  \\ 
        \cmidrule{2-4}
                                    &   GatedGCN     &  lr=1e-5, weight-decay=0, dropout=0, \#layer=4, \#hidden=64, \#epoch=1000  & $\alpha=0, \beta=0.1, \gamma=0.01$  \\ 
\midrule
        \multirow{5}{*}{PROTEINS}      &   GCN     &    lr=7e-4, weight-decay=0, dropout=0, \#layer=4, \#hidden=128, \#epoch=1000 & $\alpha=0, \beta=1, \gamma=0$  \\ 
        \cmidrule{2-4}
                                    &   GAT      &     lr=1e-3, weight-decay=0, dropout=0, \#layer=4, \#hidden=16, \#head=8, \#epoch=1000 & $\alpha=0, \beta=0, \gamma=0.01$  \\ 
        \cmidrule{2-4}
                                    &   GraphSage     &   lr=7e-5, weight-decay=0, dropout=0, \#layer=4, \#hidden=96, \#epoch=1000 & $\alpha=0, \beta=1, \gamma=1$  \\ 
        \cmidrule{2-4}
                                    &   GIN     &   lr=7e-3, weight-decay=0, dropout=0, \#layer=4, \#hidden=96, \#epoch=1000 & $\alpha=1, \beta=1, \gamma=1$  \\ 
        \cmidrule{2-4}
                                    &   GatedGCN    &   lr=1e-4, weight-decay=0, dropout=0, \#layer=4, \#hidden=64, \#epoch=1000 & $\alpha=0, \beta=1, \gamma=0$  \\ 
\bottomrule
\end{tabular}
} 
\label{tab:hypers}%\vspace{-3mm}
\end{table*}

\end{document}